\documentclass[sigconf]{acmart}

\AtBeginDocument{%
  }

\copyrightyear{2024}
\acmYear{2024}
\setcopyright{acmlicensed}\acmConference[MM '24]{Proceedings of the 32nd
ACM International Conference on Multimedia}{October 28-November 1,
2024}{Melbourne, VIC, Australia}
\acmBooktitle{Proceedings of the 32nd ACM International Conference on
Multimedia (MM '24), October 28-November 1, 2024, Melbourne, VIC, Australia}
\acmDOI{10.1145/3664647.3680849}
\acmISBN{979-8-4007-0686-8/24/10}
\usepackage{siunitx}
\usepackage{array}
\usepackage{cleveref}
\crefname{figure}{Fig.}{Figs.}
\Crefname{figure}{Fig.}{Figs.}
\begin{document}

\title{MagicFight: Personalized Martial Arts Combat Video Generation}

\author{Jiancheng Huang}
\orcid{0000-0003-2826-9231}
\authornote{Both authors contributed equally to this research.}
\affiliation{%
  \institution{Shenzhen Institute of Advanced Technology, Chinese Academy of Sciences}
  \city{Shenzhen}
  \country{China}
}
\email{jc.huang@siat.ac.cn}

\author{Mingfu Yan}
\orcid{0009-0000-4820-0704}
\authornotemark[1]
\affiliation{%
  \institution{Shenzhen Institute of Advanced Technology, Chinese Academy of Sciences}
  \city{Shenzhen}
  \country{China}
}
\email{mf.yan@siat.ac.cn}

\author{Songyan Chen}
\orcid{0009-0005-0172-6887}
\authornotemark[1]
\affiliation{%
  \institution{China Telecom Cloud Technology Co., Ltd.}
  \city{Beijing}
  \country{China}
}
\email{chensy86@mail3.sysu.edu.cn}

\author{Yi Huang}
\orcid{0000-0002-8443-6877}
\affiliation{%
  \institution{Shenzhen Institute of Advanced Technology, Chinese Academy of Sciences}
  \city{Shenzhen}
  \country{China}
}
\email{yi.huang@siat.ac.cn}

\author{Shifeng Chen}
\orcid{0000-0003-0677-7358}
\authornote{Corresponding author. This work is supported by Shenzhen Science and Technology Innovation Commission (JSGG20220831105002004).}
\affiliation{%
  \institution{Shenzhen Institute of Advanced Technology, Chinese Academy of Sciences}
  \city{Shenzhen}
  \country{China}
  }
  \affiliation{%
  \institution{Shenzhen University of Advanced Technology}
  \city{Shenzhen}
  \country{China}
}
\email{shifeng.chen@siat.ac.cn}
\renewcommand{\shortauthors}{Jiancheng Huang, Mingfu Yan,  Songyan Chen, Yi Huang, \& Shifeng Chen }

\begin{abstract}
  Amid the surge in generic text-to-video generation, the field of personalized human video generation has witnessed notable advancements, primarily concentrated on single-person scenarios. However, to our knowledge, the domain of two-person interactions, particularly in the context of martial arts combat, remains uncharted. We identify a significant gap: existing models for single-person dancing generation prove insufficient for capturing the subtleties and complexities of two engaged fighters, resulting in challenges such as identity confusion, anomalous limbs, and action mismatches. To address this, we introduce a pioneering new task, Personalized Martial Arts Combat Video Generation. Our approach, MagicFight, is specifically crafted to overcome these hurdles. Given this pioneering task, we face a lack of appropriate datasets. Thus, we generate a bespoke dataset using the game physics engine Unity, meticulously crafting a multitude of 3D characters, martial arts moves, and scenes designed to represent the diversity of combat. MagicFight refines and adapts existing models and strategies to generate high-fidelity two-person combat videos that maintain individual identities and ensure seamless, coherent action sequences, thereby laying the groundwork for future innovations in the realm of interactive video content creation.

  \noindent\textbf{Website:} \url{https://MingfuYAN.github.io/MagicFight/}

  \noindent\textbf{Dataset:} \url{https://huggingface.co/datasets/MingfuYAN/KungFu-Fiesta}
\end{abstract}

\vspace{-6pt}

\vspace{-4pt}

\begin{teaserfigure}
  \includegraphics[width=\textwidth]{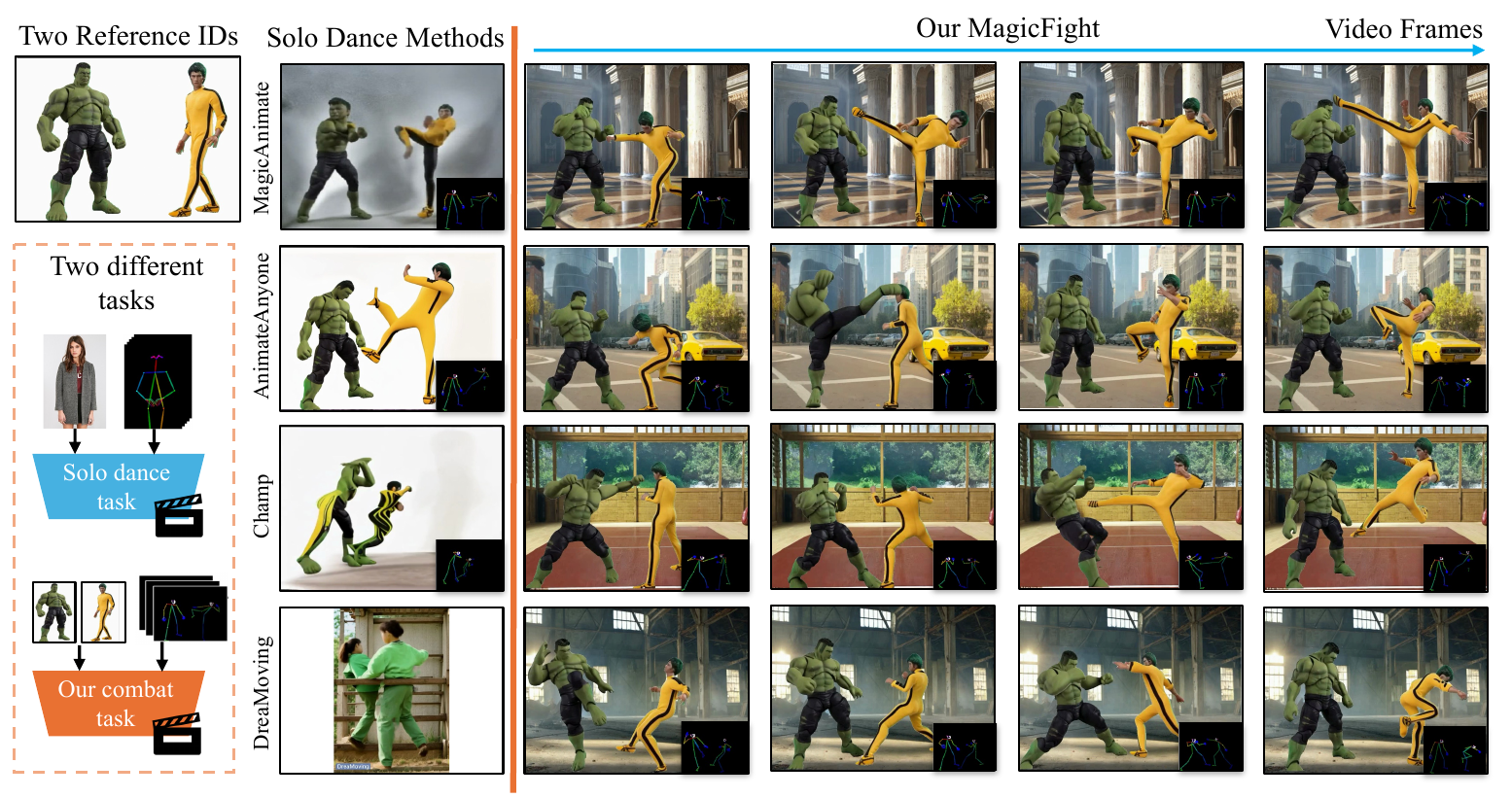}
  \caption{Our method is the first method capable of generating high-quality martial arts combat videos. It takes two reference ID images and a conditioned pose sequence as input and generates a video that maintains consistency in both IDs and action. The solo dance methods struggle with this new task. The right part showcases our results.}
  \Description{}
  \label{fig:teaser}
\end{teaserfigure}

\maketitle

\vspace{-9pt}
\section{Introduction}
Video generation has emerged as a prominent field in AI research in recent years, with the creation of personalized videos representing a subtask of significant commercial and artistic value. When the specified subject is a human, this process, also known as character animation generation, entails providing an image of the source character, whereupon the model generates a realistic video following a sequence of poses specified by the user. This task boasts many potential applications, including online retail, entertainment videos, art creation, and virtual characters, among others. Numerous studies have explored image animation and pose transfer by GAN\cite{fomm,mraa,ren2020deep,tpsmm,siarohin2019animating,zhang2022exploring, bidirectionally,everybody}, serving as foundational work.

In recent years, diffusion models~\cite{denoising,huang2023bootstrap,huang2023wavedm} have demonstrated their superiority in text-to-image~\cite{dalle2,glide,imagen,ldm,composer,ediffi,liao2024freehand,huang2024survey,huang2024bk,huang2024entwined,huang2024sbcr,chen2023specref} and video generation~\cite{animatediff,cogvideo,fatezero,imagenvideo,text2videozero,tuneavideo,videocomposer,align,gen1,makeavideo,vdm}. Numerous researchers have utilized the architecture of diffusion models to explore video generation conditioned on given image~\cite{videocomposer,videocrafter1,i2vgen,animatediff}. However, when applied to human animation, for which they are not specifically designed, these methods often produce character appearances that do not match the original image, leading to videos that lack movement coherence. For fashion video generation, DreamPose\cite{dreampose} introduces an adapter to fuse CLIP\cite{clip} image features into Stable Diffusion~\cite{ldm} and finetunes on the input sample.

Recent works specializing in human dance video generation, including DisCo~\cite{disco}, MagicAnimate~\cite{xu2023magicanimate}, AnimateAnyone~\cite{hu2023animate}, MagicDance~\cite{chang2023magicdance}, DreaMoving~\cite{feng2023dreamoving} and Champ~\cite{zhu2024champ} exhibit similar approaches and network structures. DisCo~\cite{disco} extracts character and background features via ControlNet\cite{controlnet} while it shows serious flaws in generating the ID. Other methods~\cite{xu2023magicanimate,hu2023animate,chang2023magicdance,feng2023dreamoving,zhu2024champ} all aim to solve the issue of ID appearance. Each employs its own appearance encoder, utilizing a parameter-rich encoder like ControlNet for multi-scale and detailed ID feature extraction from the original image. They design an effective pose guide for controllability and a temporal module for smooth interframe transitions. Furthermore, the pivotal element is the training data they have amassed. By leveraging large-scale, high-quality datasets, these methods can animate arbitrary characters.

However, all the aforementioned methods fall short in human fighting video generation involving multiple subjects. As these methods are designed for single-person dancing, they accept a single ID and a single-person pose sequence, and their training datasets predominantly contain single-person dance videos such as TikTok~\cite{tiktok}. Besides, the absence of network design for multi-person and the lack of multi-person dataset preclude these existing works from effectively generating multi-person fighting videos. Hence, we introduce a new task: personalized martial arts combat video generation. There are three primary distinctions between our new task and the existing ones: 1) Subject number: The existing task focuses on solo dances, whereas ours involves two individuals. 2) Motion type: While fashion and dance videos emphasize slow and individual movements, martial arts combat requires capturing complex kung fu and varied poses. 3) Interaction dynamics: Unlike solo dance with no interactive dynamics, martial arts combat necessitates depicting the intricate interplay between two-person, highlighting the authenticity of the generated video.

In this paper, we design a base method MagicFight for our proposed new task named personalized martial arts combat video generation. To establish this foundational method, we address existing issues in current techniques and investigate dataset production, processing, and training strategies. Our main contributions include:

\vspace{-5pt}

\begin{enumerate}

\item For the first time, we delineate two-person fighting from one-person dancing. We create a dataset of martial arts combat videos named KungFu-Fiesta (KFF) and establish data cleaning rules for dataset quality and diversity, laying a solid foundation for this new task.

\item We introduce a multi-modal personalized network to learn conditioning on two reference IDs, pose, background, and prompt, focusing on the dynamic complexities of combat. With the personalized attention layer (ID-attn), we address the clothing and body misattribution problem in our task.

\item In the inference stage, we introduce body-shape adaptive strategy to automatically adjust the preset pose map, aligning the generated video more closely with the expected body shape. For arbitrary long video generation, we use a clip fusion technique to ensure continuity between clips.

\item We conduct a comprehensive ablation study from both the dataset and model training perspectives. We explore the properties, size, and quality requirements for our dataset on this task,  and offer insights and guidance about the effect of different training components on overall performance. We creatively propose the Mixture Data Finetuning strategy, which mixes self-made two-person fashion videos and KFF dataset for training, in order to take full advantage of different data domain.

\end{enumerate}

\vspace{-8pt}
 \section {Related Work}
\subsection{Conditional Video Generation}
The field of video generation has advanced significantly, thanks to diffusion models adapted from text-to-image (T2I) techniques. Research efforts \cite{text2videozero,fatezero,cogvideo,tuneavideo,rerender,gen1,followyourpose,makeavideo,vdm} introduce frame attention and embedding temporal layers within T2I models. Initiatives like Video LDM \cite{align} advocate for image pretraining before engaging in video temporal training, and AnimateDiff \cite{animatediff} brings motion modules to T2I models without the need for specialized adaptation. Expanding into image-to-video transformation, VideoComposer \cite{videocomposer} stands out by integrating images as conditional inputs during training. VideoCrafter \cite{videocrafter1} distinguishes itself by melding text and visual features from CLIP into its cross-attention mechanism. The Stable Video Diffusion (SVD)~\cite{blattmann2023stable} signifies a quantum leap in enhancing video quality and dynamic representation. With W.A.L.T~\cite{gupta2023photorealistic} pioneering through its VAE Encoder in choosing optimal latent representations, and the Sora~\cite{sora} setting new standards for high-definition, realistic video outputs, these advancements mark a decisive turn towards refined, high-quality video generation.
\vspace{-8pt}
\subsection{Human Video Generation}
Recent studies \cite{gong2023tm2d,he2024co,hu2024structldm,xu2024you} highlight the incorporation image-to-video diffusion model into human video generation. PIDM\cite{pidm} introduces texture diffusion blocks to infuse desired texture patterns into the denoising process for human pose migration. LFDM\cite{LFDM} synthesizes optical flow sequences in latent space, distorting the input image based on specified conditions. LEO\cite{leo} represents motion through a series of flow maps, using a diffusion model to synthesize the motion sequence. DreamPose\cite{dreampose} utilizes a pre-trained stable diffusion model, introducing an adapter to model the image embeddings extracted by CLIP and VAE. DisCo\cite{disco}, inspired by ControlNet, decouples pose and background control. MagicAnimate, AniamteAnyone, and Champ build primarily on the DisCo and advance the improvement of appearance alignment and motion control mechanisms. However, these methods still struggle with issues like ID appearance inconsistency and temporal instability. Moreover, no method yet exists for generating martial arts combat videos or focusing on two-person motion video.

\begin{figure*}[t]
\begin{center}
	\setlength{\fboxrule}{0pt}
	\fbox{\includegraphics[width=0.98\textwidth]{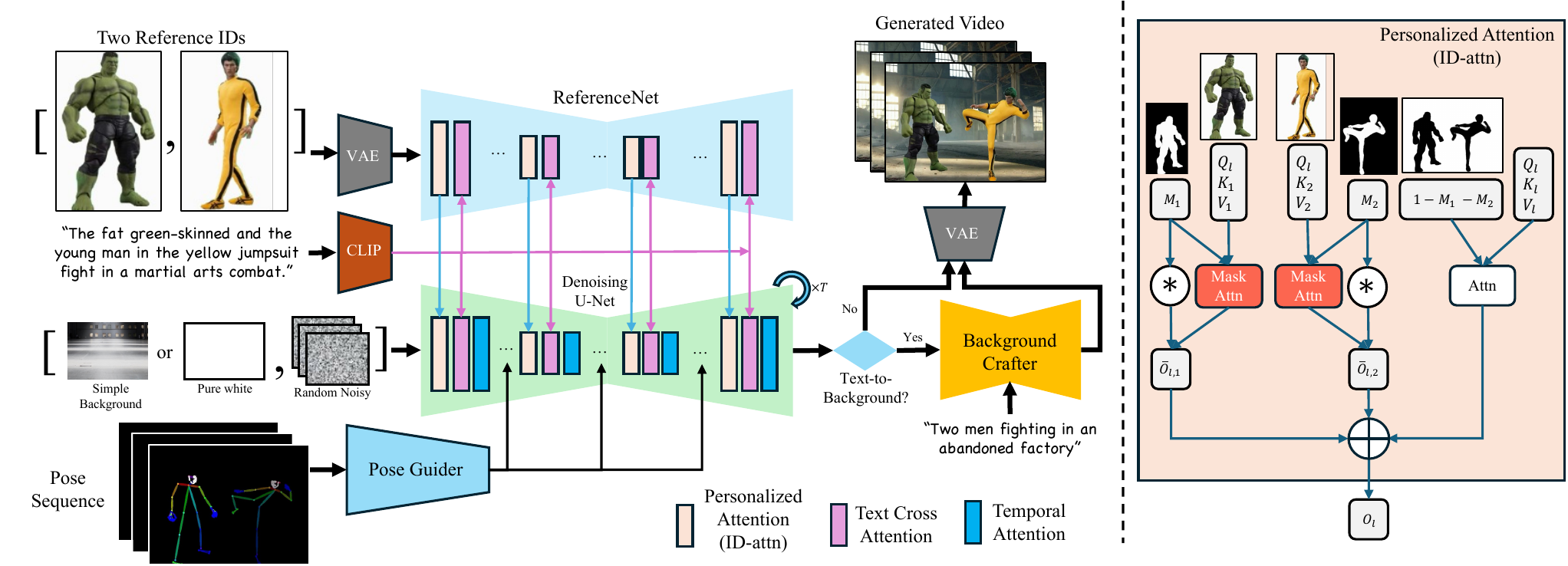}}
\end{center}
\vspace{-0.3cm}
\caption[Framework for duo martial arts video generation]{
Our MagicFight has 4 conditions for combat video generation, two reference IDs, a text prompt, a background image, and pose maps. The action of each frame is controlled by Pose Guider. The two IDs are personalized by our Personalized Attention (ID-attn) layer which can generate the respective appearance to the desired place. The user can provide a simple background image or use a pure white background and then generate a complex and reasonable background by our Background Crafter. With the long video generation technique, we can make arbitrary long videos (typically 10 seconds in our test).
}
\label{fig:video_overview}
\vspace{-0.4cm}
\end{figure*}

\vspace{-5pt}
 \section{Methods}

First, we analyze the existing problems in Sec.~\ref{sec:motivation}. Then, we detail our dataset creation process in Sec.~\ref{sec:video_dataset}, and our model architecture in Sec.~\ref{sec:model}. We describe the training
 and inference in Sec.~\ref{sec:train} and~\ref{sec:infer}, respectively.

 \begin{table}[t]
\centering
\caption{Details of Our Martial Arts Combat Video Dataset}
  \vspace{-0.2cm}
\label{tab:dataset-details}
\vspace{-3pt}
\resizebox{1\linewidth}{!}{
\begin{tabular}{lS[table-format=3.0]S[table-format=3.0]S[table-format=3.0]S[table-format=3.0]S[table-format=3.0]S[table-format=3.0]}
\toprule
\textbf{Scene} & {City} & {Palace} & {Desert} & {Seaside} & {Mountains} & {Snowfield} \\
\textbf{No. of Videos} & 41 & 45 & 43 & 40 & 39 & 45 \\
\textbf{} & {Island} & {Bridge} & {School} & {Rainforest} & {Open Field} & {Boxing Gym} \\
\textbf{No. of Videos} & 42 & 37 & 36 & 38 & 44 & 35 \\
\midrule
\textbf{Action} & {Judo} & {Dodging} & {Blocking} & {Wing Chun} & {Jeet Kune Do} & {Weapon Combat} \\
\textbf{No. of Videos}  & 42 & 36 & 45 & 41 & 39 & 44 \\
\textbf{} & {Boxing} & {Kicking} & {Wrestling} & {Somersault} & {Rapid Attack} & {Finishing Move} \\
\textbf{No. of Videos}  & 40 & 42 & 43 & 38 & 37 & 34 \\
\midrule
\textbf{Character} & {Beauty} & {Athlete} & {Soldier} & {Male Staff} & {Thin Person} & {Short Person} \\
\textbf{No. of Videos}  & 40 & 42 & 45 & 43 & 42 & 39 \\
\textbf{} & {Elderly} & {Student} & {Fat Person} & {Tall Person} & {Female Staff} & {Martial Artist} \\
\textbf{No. of Videos}  & 36 & 34 & 38 & 37 & 41 & 44 \\
\bottomrule
\end{tabular}
}
\vspace{-0.60cm}
\end{table}

\vspace{-5pt} 
 \subsection{Existing Problems and Motivation}\label{sec:motivation}

We commence with an analysis of the challenges that existing models face in generating scenes with complex character interactions as shown in Fig.~\ref{fig:teaser}. 1) During the generation with two-person interactions, a common issue is misattribution of clothing and body parts, particularly when characters are close. For instance, the woman’s left leg in the short skirt might be incorrectly merged with the man’s pants, with color inaccuracies also occurring. These issues highlight the necessity for a customed two-person model to address the misattribution problem. 2) Besides, missing body parts also frequently occur, like duplicated legs or absent feet. This issue stems from the inadequate data on leg lifting and kicking actions in the human dance dataset and the absence of foot keypoints in the pose maps, leading to the model's poor perception of leg and foot features, which is thirsty for a tailored martial arts dataset. 3) Moreover, existing models often produce medium-sized characters, overlooking the diversity in body shape. For instance, muscular "Hulk" and bony people are frequently underrepresented. Hence, we aim to solve the problem of mismatch between body type and given pose, adapting to any body shape during inference. Our research seeks to mitigate the aforementioned problems and lay the groundwork for future endeavors.

\vspace{-6pt}
\subsection{KungFu-Fiesta Dataset Creation}\label{sec:video_dataset}
\vspace{-1pt}
This section details our first martial arts combat video dataset named KungFu-Fiesta (short for KFF). We make 4 scenarios for this dataset creation and finally chose the Unity scenario, details about it are in our appendix. With Unity, an advanced game physics engine, it can create highly realistic 3D character models and action animation in a simulated world, and by rendering the scene from an angle and exporting them to video, it is possible to create a large number of highly realistic martial arts combat videos. For the diversity and complexity of the dataset, we design hundreds of character IDs with different identities, covering more than 100 kinds of paired fighting actions, and a variety of shooting angles in 20 different scenes. After careful design and production, we capture more than 500 high-quality videos. Each video sample is about 10 seconds with 60 fps, ensuring the coherence of the action. In KungFu-Fiesta, each sample contains a combat video, two reference images of character IDs (for short reference IDs), and a pose map sequence, providing researchers with more conditions. The details of the dataset are shown in Table~\ref{tab:dataset-details}.

\vspace{-6pt}
\subsection{Multi-Modal Personalized Network}\label{sec:model}
\vspace{-1pt}
Our model is an extension of the Stable Diffusion (SD), so we inherit its VAE \cite{vae}, denoising UNet and CLIP encoder.
Fig.~\ref{fig:video_overview} provides an overview of our framework. The input to the network is multi-frame noisy latent $z_t \in {\mathbb R}^{{\mathnormal F}{\times}{\mathnormal c}{\times}{\mathnormal h}{\times}{\mathnormal w}}$ (timestep $t$). In order to utilize the general knowledge of human motion, our model is based on the pretrained model of AnimateAnyone~\cite{hu2023animate} which is for single-person dancing video generation. The framework consists of three key components: 1) ReferenceNet is responsible for encoding the appearance of the two reference IDs; 2) Pose Guider is for controlling the two-person's fight by pose map; 3) Temporal layer, the attention layer between these frames is to ensure the continuity of the character's movement. AnimateAnyone proposes to use a lightweight pose controller with only 4 simple convolutional layers since the pose control of single-person is easy. However, our two-person martial arts situation is more complex and the input pose becomes two-person. We finally choose to use a large Pose Guider like ControlNet.

\textbf{Personalized Attention Layer.}
In our task, the given reference IDs provide detailed appearance information. However, the ReferenceNet of AnimateAnyone is designed and trained for single-person feature extraction. Thus, we feed 2 ID images into ReferenceNet alongside the batchsize dimension to extract their features $[r_1, r_2]$, and then they are fed into the denoising U-Net. As shown in Fig.~\ref{fig:video_overview}, our personalized attention (ID-attn) layer replaces the original self-attention layer of SD. Given the feature map ${x}_{l} {\in} {\mathbb R}^{{\mathnormal F}{\times}{\mathnormal h}{\times}{\mathnormal w}{\times}{\mathnormal c}}$ in the $l$-th ID-attn layer and ID features $r_1, r_2 \in {\mathbb R}^{{\mathnormal h}{\times}{\mathnormal w}{\times}{\mathnormal c}}$, ID-attn is performed as:
\vspace{-0.5cm}

\begin{equation}
 \vspace{-0.18cm}
 \label{eq:mask-attn}
 \begin{aligned}
 \bar O_{l,i} =&\text{MaskAttn}(Q_{l}, K_{i},M_i)V_{i},\\
 O_{l} = &M_1 \bar O_{l,1}  +M_2 \bar O_{l,2} +(1-M_1-M_2) \text{Attn}(Q_{l}, K_{l})V_{l},
 \end{aligned}
\end{equation}

where $Q_l$ denotes the Query from $x_l$, $[K_i, V_i]$ represents the Key/Value from $i$-th ID $r_i$, and $[K_l, V_l]$ denotes the Key/Value from $x_l$ itself. Computing attention between the whole $x_l$ and $r_i$ may lead to reference disruption. Thus, $\text{MaskAttn}()$ means only to keep the attention of the $M_i$ region and mask the other regions with no attention. $M_i$ denotes the target position of ID $i$, computed by the bounding box of the pose of ID $i$. So the target part of $M_i$ is from $r_i$ and the background part $1-M_1-M_2$ is not affected by IDs.

\textbf{Conditioned Background.}
For conditioned background (pure white also OK), we concat the given background image latent with $z_t$ at channels and input to U-Net. The user can 1) provide a simple background image for end-to-end background customisation, and 2) if the user does not want to provide a background image, the conditioned background will be set to pure white, and then user can provide text prompt in Background Crafter to generate the background. Our Background Crafter is based on SDXL-Inpainting~\cite{podell2023sdxl}. Its conditions are the original foreground image, background mask (it is easy to obtain a mask due to the white background), and text prompt. We finetune it on our dataset and follow \cite{lv2023gpt4motion} to maintain inter-frame background consistency, which is detailed in our appendix.

\vspace{-6pt}
 \subsection{Multi-Stage and Mixture Dataset Finetuning}\label{sec:train}

 \subsubsection{Mixture Dataset Finetuning}

We propose to use a mixture of KFF dataset with our recreated two-person fashion video dataset made based on the UBC dataset ~\cite{dwnet}. It is worth noting that the UBC dataset originally contains videos of a single person walking down a fashion runway, and by splicing two randomly selected videos left and right together, we create a video dataset that simulates a two-person fashion runway, which has the advantages of a pure white and clean background, real people in the subjects, and high-definition clothing textures, and the disadvantages that the two subjects are randomly spliced together, and lack of multi-subject interactions. Based on the benefits of the two-person fashion dataset, we hypothesise that a strategy of training with a mixture of two-person fight videos and two-person fashion videos would improve the consistency and aesthetics of the appearance, thus demonstrating stronger generalisation capabilities when dealing with complex character interaction scenarios.

\vspace{-4pt}
 \subsubsection{Multi-Stage Finetuning}
Since our MagicFight model is finetuned on the pretrained Moore-AnimateAnyone~\cite{moore2024animate}\footnote{Since AnimateAnyone has no released code, we use reproduction version.}, we propose the 2 stages finetuning. The first stage during finetuning is a spatial learning stage, using individual video frames from our KFF dataset as image input. In denoising U-Net, all temporal layers are frozen and become intra-frame attention, and the model takes the noisy single frame as input, along with the reference IDs and the pose map. ReferenceNet and Pose Guider are trained to learn the spatial distribution of the two-person fighting. The pretrained weights on the human dancing dataset are used to initialize our denoising U-Net, ReferenceNet, and Pose Guider, which adopts a ControlNet-like structure rather than the lightweight controllers in Animate Anyone. The second stage is for temporal layer finetuning, whose input is 20 frames of video clip from our KFF dataset, and network parameters except temporal layers are frozen to learn the general law of two-person fighting action.

\vspace{-6pt}
 \subsection{Inference Strategy}\label{sec:infer}

\subsubsection{Long Video Generation Technique}
Previous diffusion-based video generation typically focuses on short video clips. For generating an arbitrary long combat video, we introduce a clip fusion technique to ensure continuity of details between clips. Specifically, we retain the $x_t$ of the last 4 frames of each clip in sampling steps. When inferring the next video clip, we use the saved 4 frames and the following 20 frames as $x_t$. During each sampling step, we superimpose the $x_t$ of the last 4 frames onto those of the first 4 frames of the currently generated clip to generate videos of arbitrary length maintaining consistency.

 \begin{figure}[t]
\begin{center}
	\setlength{\fboxrule}{0pt}
\fbox{\includegraphics[width=\linewidth]{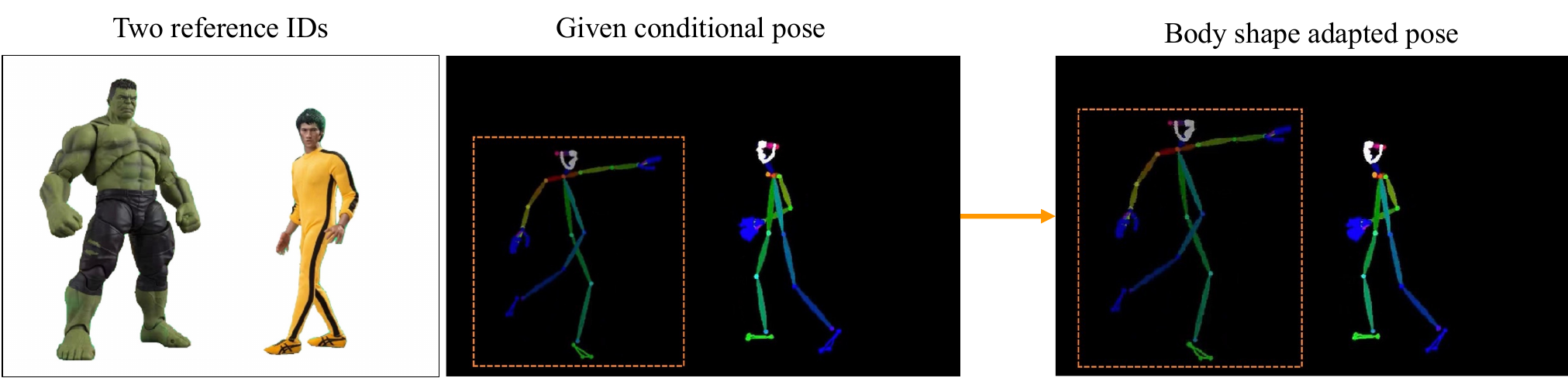}}
\end{center}
\vspace{-0.3cm}
\caption{The body size adaptive strategy during inference.}
\Description{}
\label{fig5:body_adp}
\vspace{-0.5cm}
\end{figure}

 \subsubsection{Body-Shape Adaptive Strategy}

 In the video generation process, considering the possible differences in body types (e.g., height, body shape, etc.) between the given pose map and the reference IDs, we face a challenge to ensure the body shape in the generated video is consistent with those in the reference IDs. For example, if the reference ID is a tall and chubby person, and the given pose is from a little girl, it may result in visual incongruity. Thus, we introduce a body-shape adaptive strategy, which is shown in Fig.~\ref{fig5:body_adp}. First, we predict the pose of the reference IDs, and compute the center of mass of the character's keypoints in the horizontal (x-axis) and vertical (y-axis) directions. Similarly, we also compute those of the given conditioned pose map. Subsequently, we compute the body scale factors in the x/y-axis. With these scale factors, we scale the coordinates of all the key points in the pose to ensure that the generated video content meets the action requirements and is faithful to the body shape of the reference IDs.

\vspace{-6pt}
\section{Experiment}

\vspace{-2pt}

\subsection{Implementation}

To validate MagicFight's efficacy in generating martial arts combat videos with diverse IDs, we make two benchmarks, KFF reconstruction benchmark and open-set combat generation benchmark, to evaluate our model. We employ pretrained DWPose to estimate pose maps, including body, hands, and foot. All finetuning experiments are conducted on 8 NVIDIA A6000 GPUs, each with 48G GPU memory. In the first finetuning stage, we sample individual frames at a frame interval of 6, then adjust the frames to a resolution of $704{\times}512$. Finetuning is performed for 20,000 steps, with a batchsize of 2 per GPU. In the second finetuning stage, we finetune the temporal layer for 10,000 steps with a video sequence of 20 frames, frame interval of 6, and batchsize of 2. Both learning rates are set to 2e-6. During inference, we employ the DDIM sampling for 25 denoising steps. We adopt our long video generation technique and body-shape adaptive strategy for better generation. For comparison with human dance generation methods, we test all methods on the same benchmark, detailed in Sec.~\ref{com}.

 \begin{figure*}[t] \begin{center}
 \setlength{\fboxrule}{0pt}
 \setlength{\fboxsep}{0pt}
 \fbox{\includegraphics[width=0.95\linewidth]{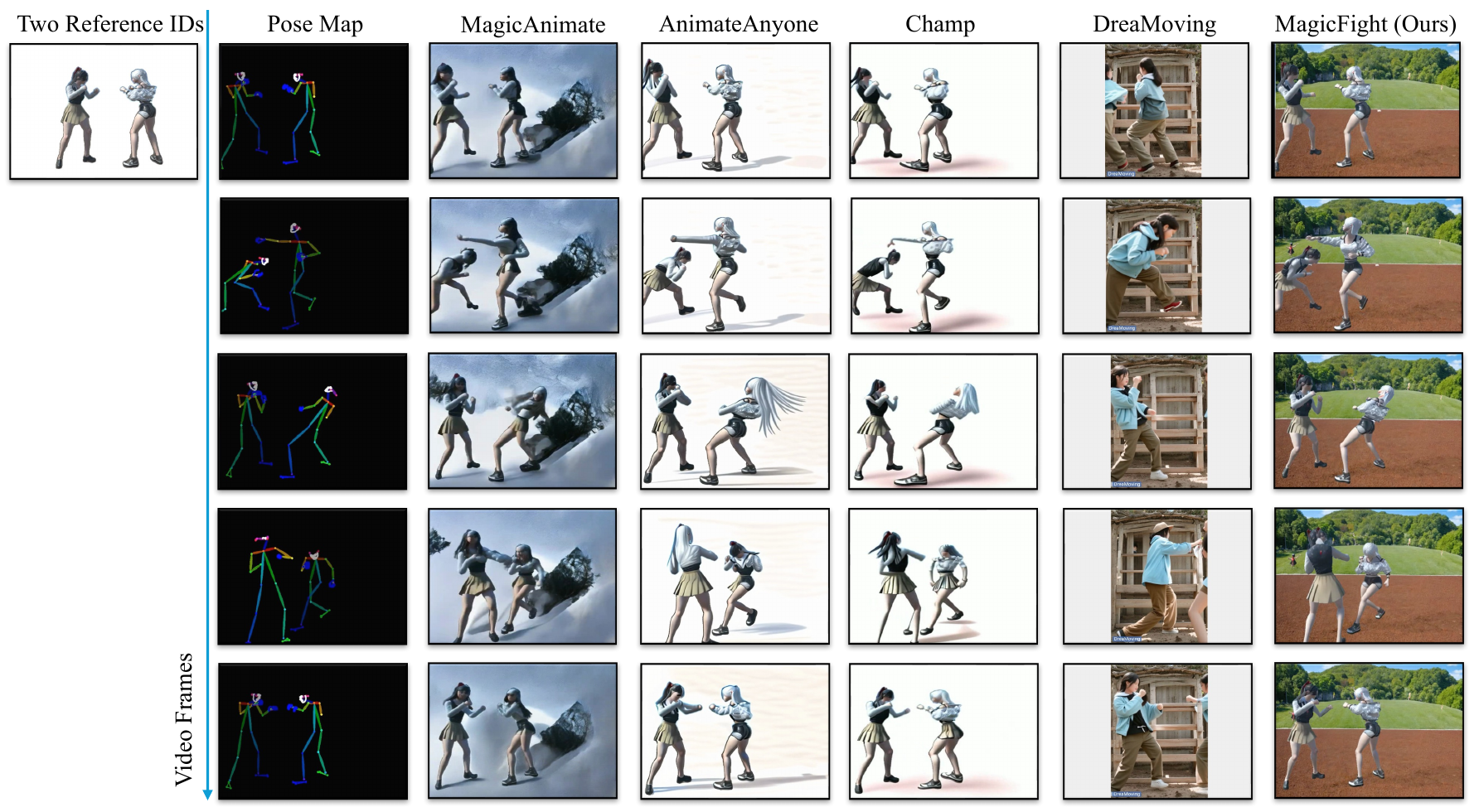}}
 \fbox{\includegraphics[width=0.95\linewidth]{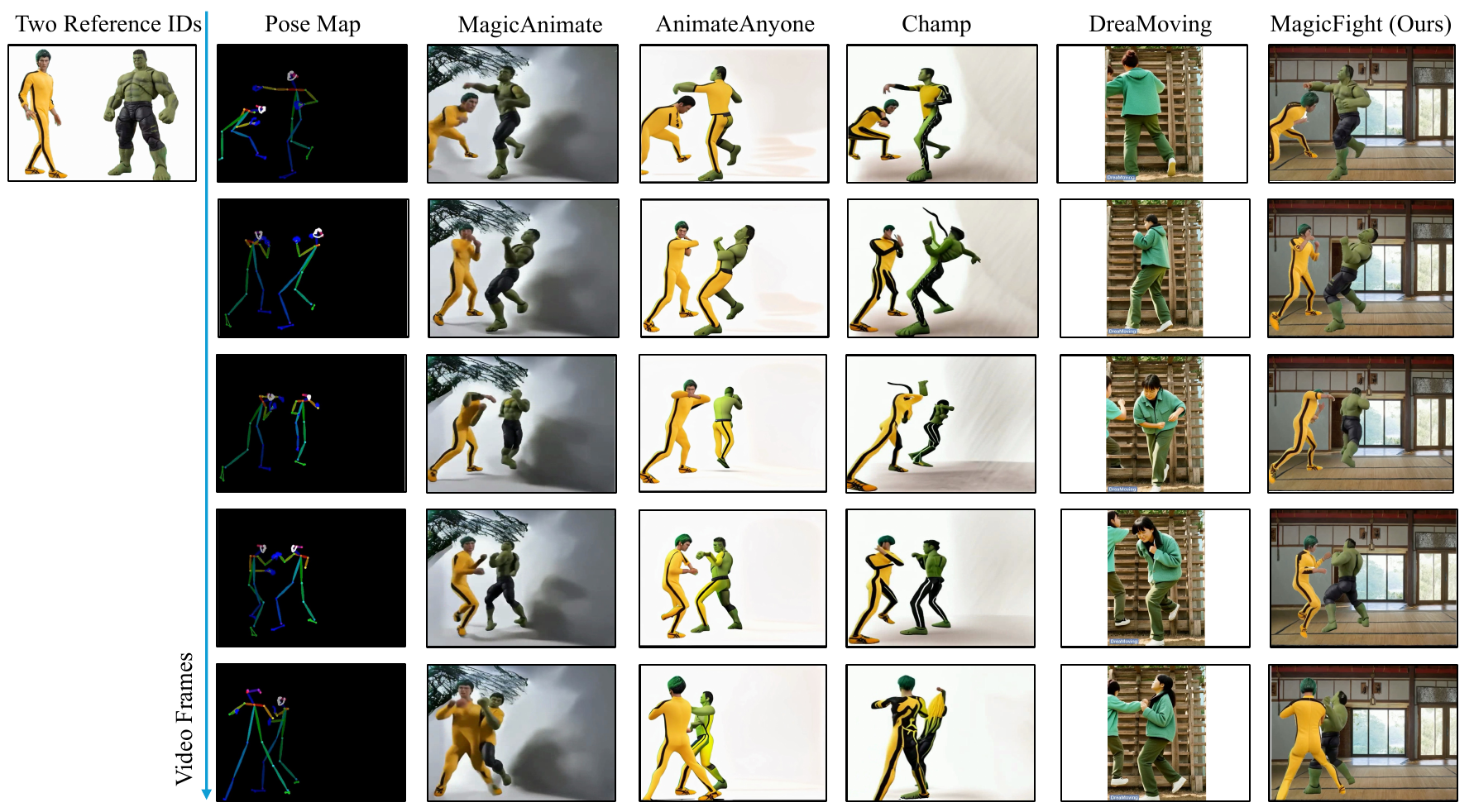}} \end{center}
  \vspace{-0.4cm}
  \caption[Results of the two-person martial arts model on the test set 3]{
 The results on two benchmarks. These solo dance models exhibit missing body parts and wrong actions, and they cannot be conditioned on background or generate background by prompt. Our MagicFight significantly mitigates these issues.
 }
 \Description{}

 \label{fig5:edit1}
\vspace{-0.5cm}
 \end{figure*}

\begin{figure*}[t] \begin{center} \setlength{\fboxrule}{0pt} \fbox{\includegraphics[width=0.965\linewidth]{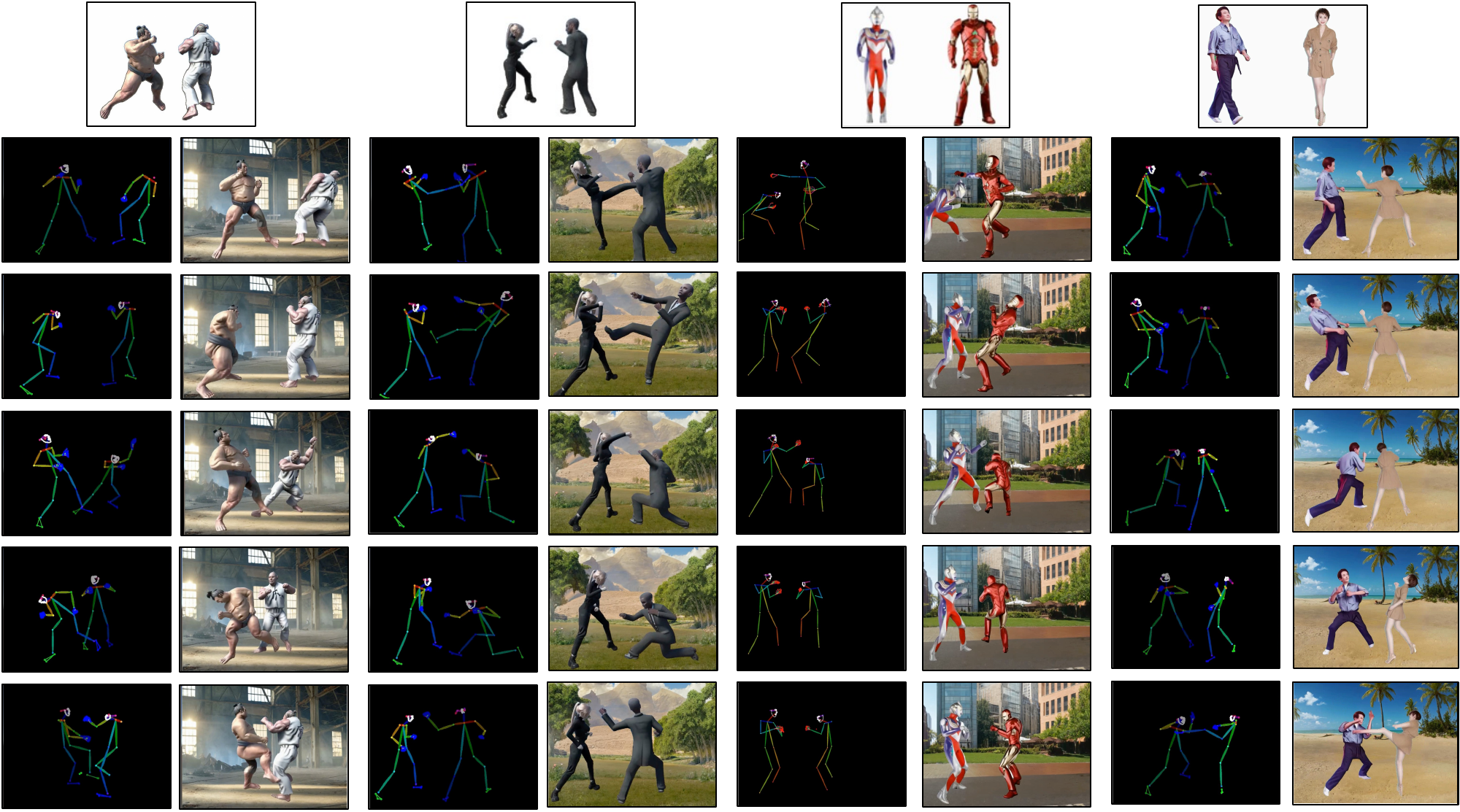}} \end{center}
  \vspace{-0.4cm}
\caption[Results of the two-person martial arts model on the test set 2]{
 The MagicFight results in open-set combat generation with smooth movements and consistent IDs. Because of page limits, we give more results in our appendix.}
 \Description{}
   \vspace{-0.4cm}
 \label{fig5:edit4} \end{figure*}

\vspace{-9pt}
  \subsection{Qualitative and Quantitative Evaluation}\label{com}

\Cref{fig5:edit1,fig5:edit4} illustrate our method's capability to produce controllable combat videos for various character types, such as real, cartoon, robotic, and humanoid. Our method produces high-definition videos with realistic character details. It ensures temporal consistency with the reference IDs and maintains continuity between frames, despite significant motion.

To illustrate our method's superiority over other video generation methods, we assess them on two bespoke benchmarks: KFF reconstruction benchmark and open-set combat generation benchmark. For quantitative evaluation of the reconstructed video quality, we utilize metrics such as SSIM\cite{ssim}, PSNR\cite{psnr}, FVD\cite{fvd}, and LPIPS\cite{lpips}. The evaluation of the open-set video generation benchmark incorporates user ratings, FVD\cite{fvd}, and NIQE metrics\cite{mittal2012making}. In our experiments, we follow the computation of FVD as VideoGPT~\cite{yan2021videogpt}.

Given that SSIM and PSNR may not match human perception, we employ LPIPS and NIQE as complementary evaluation metrics. LPIPS quantifies perceptual similarity, offering a closer representation of the human eye's subjective judgment. NIQE acts as a reference-free image quality evaluation metric tailored to appraise the visual aesthetic quality of images. A user study is conducted to evaluate the subjective quality comprehensively. Forty users review the results from all methods. Each sample consists of IDs image, pose sequence, text prompt, and results from each method. Participants rate the quality of each video on a scale from 1 to 5. The evaluation primarily focuses on the IDs’ similarity, pose control and visual appeal. We calculate the average scores for each method and gauge potential popularity and practical value.
\begin{table}[t]
	\centering
	\caption{Quantitative comparison of two benchmarks.}
	\vspace{-10pt}
	\label{table:combined}
	\resizebox{\linewidth}{!}{
 \begin{tabular}{lcccccccc} 
		\toprule
		\multicolumn{4}{c}{Open-set Combat Generation Benchmark} & \multicolumn{4}{c}{KFF Reconstruction Benchmark} \\
		\cmidrule(lr){1-4} \cmidrule(lr){5-8}
		Method & FVD $\downarrow$ & NIQE $\downarrow$ & User Score $\uparrow$ & SSIM $\uparrow$ & PSNR $\uparrow$ & LPIPS $\downarrow$ & FVD $\downarrow$ \\ 
		\midrule
		MagicAnimate & 937.34 & 5.23 & 2.05 & 0.888 & 22.479 & 0.090 & 623.00 \\
		AnimateAnyone & 1178.57 & 4.68 & 3.77 & 0.873 & 21.398 & 0.087 & 572.22 \\
		Champ & 1130.22 & 4.56 & 3.89 & 0.877 & 22.018 & 0.066 & 523.01 \\
		DreaMoving & 1851.93 & 5.92 & 0.41 & --- & --- & --- & --- \\ 
		Ours & \textbf{812.77} & \textbf{4.14} & \textbf{4.12} & \textbf{0.893} & \textbf{23.756} & \textbf{0.058} & \textbf{454.62} \\
		\bottomrule
	\end{tabular}}
	\vspace{-0.53cm}
\end{table}
\vspace{-4pt}
\subsubsection{The KFF Reconstruction Benchmark}
KFF reconstruction involves generating a reconstructed video given two reference IDs and the pose sequence. Our KFF reconstruction benchmark comprises 100 video clips, each with around 180 frames. The selection criteria for this benchmark require the test video's character and action to match the training set's domain, yet not exactly existing in the training set. Quantitative comparisons are detailed in Table~\ref{table:combined}, where our results significantly surpass those of other methods, particularly in the reconstruction metrics. Qualitative comparisons are displayed in Fig. \ref{fig5:edit1}. We employ the web demo or the code of the compared methods. These methods can't provide conditional backgrounds or generate new backgrounds, and DreaMoving is a vertical screen resolution (so we keep it vertical). Refining human details demands high precision, while our method maintains detail consistency.

\vspace{-4pt}
\subsubsection{Open-Set Combat Generation Benchmark}
The open-set combat generation benchmark focuses on the open world of human interactions video. We collect 20 IDs from the game community and the Internet, comprising 40 test samples. We generate about 10 seconds of video for each sample. The selection criteria for this benchmark allow characters, actions, and backgrounds to span any data domain, with no restrictions on data source. Our approach yields the best quantitative results, as shown in Table~\ref{table:combined}. DisCo, AnimateAnyone, and MagicAnimate undergo extensive pre-training on human image datasets, learning basic single-person patterns, thus lacking multi-person interaction knowledge. In contrast, our mixture dataset training on the KFF and two-person fashion dataset yields superior results compared to these methods. Our method demonstrates that without explicit segmentation, the model can discern foreground-background relationships from multi-subject movements. Furthermore, our model excels at maintaining visual continuity in complex action sequences, demonstrating robustness in handling varied character appearances.

\begin{table}[t]
    \centering
    \caption{Quantitative Comparison of Ablation Study (Note: \# represents "Number of".)}
      \vspace{-0.28cm}
    \label{table:ablation}
    \resizebox{\linewidth}{!}{
    \begin{tabular}{lcccccc}
    \hline
    Setting & \# Videos & \# IDs & \# Actions & \# Backgrounds & FVD $\downarrow$ & User Score $\uparrow$ \\
    \hline
    Small-scale & 10 & 16 & 24 & 2 & 948.45 & 3.29 \\
    Medium-scale & 50 & 60 & 80 & 4 & 880.68 & 3.98 \\
    Large-scale & 160 & 180 & 120 & 4 & \textbf{812.77} & \textbf{4.25} \\
    \hline
    Few IDs & 40 & 8 & 80 & 2 & 1012.83 & 2.56 \\
    Many IDs & 40 & 70 & 80 & 2 & \textbf{867.43} & \textbf{4.11} \\
    \hline
    Few Actions & 40 & 30 & 24 & 2 & 885.62 & 3.47 \\
    Many Actions & 40 & 30 & 80 & 2 & \textbf{848.14} & \textbf{4.09} \\
    \hline
    Single Background & 100 & 50 & 70 & 2 & \textbf{816.46} & \textbf{4.23} \\
    Various Backgrounds & 100 & 50 & 70 & 12 & 923.19 & 2.93 \\
    \hline
    Freeze Denoising U-Net & 160 & 180 & 120 & 4 & 908.64 & 3.22 \\
    Freeze ReferenceNet & 160 & 180 & 120 & 4 & 838.14 & 3.91 \\
    Finetune Pose Guider Only & 160 & 180 & 120 & 4 & 925.83 & 2.73 \\
    Finetune ReferenceNet Only & 160 & 180 & 120 & 4 & 913.92 & 2.84 \\
    Finetune Denoising U-Net Only & 160 & 180 & 120 & 4 & 823.91 & 4.01 \\
    Finetune Temporal Layer Only & 160 & 180 & 120 & 4 & 846.70 & 3.57 \\
    Train the Entire Network & 160 & 180 & 120 & 4 & \textbf{812.77} & \textbf{4.25} \\
    \hline
    Incremental Data Finetuning & 50+110 & 60+120 & 80+40 & 4 & 821.07 & 4.11 \\
    Full Data Finetuning & 160 & 180 & 120 & 4 & \textbf{812.77} & \textbf{4.25} \\
    \hline
    Only KFF Dataset & 160 & 180 & 120 & 4 & 812.77 & 4.25 \\
    Mixture Dataset & 600 & 500 & 400 & 10 & \textbf{756.43} & \textbf{4.78} \\
    \hline
    \end{tabular}
    }
  \vspace{-0.55cm}
\end{table}

\begin{figure*}[t]
\begin{center} \setlength{\fboxrule}{0pt} \fbox{\includegraphics[width=0.9\linewidth]{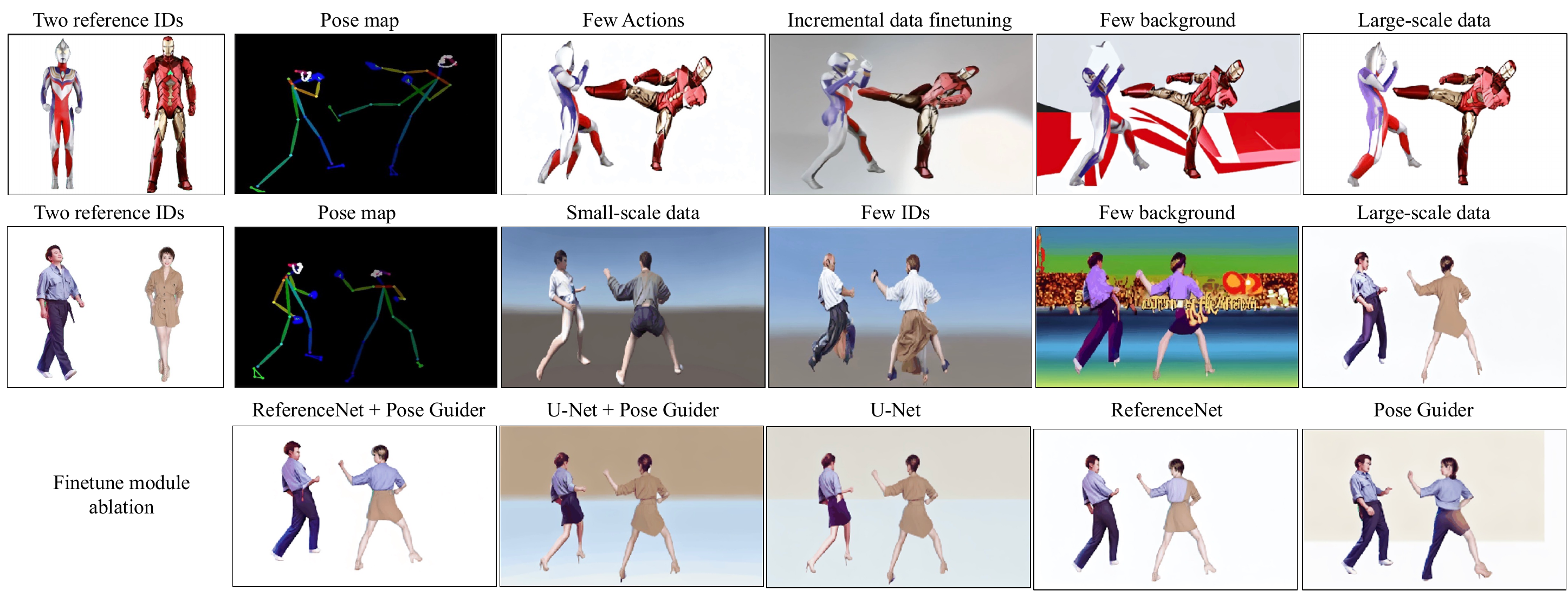}
}
\end{center}
  \vspace{-0.4cm}
\caption[Data Ablation Experiment]{
Ablation Study.
1) ID appearance is ensured with sufficient IDs in the training data. 2) Adequate action in the training set is helpful. 3) Currently, end-to-end way struggles to handle complex backgrounds.
}
\Description{}
\label{fig5:ablation}
  \vspace{-0.4cm}
\end{figure*}

\vspace{-5pt}
\subsection{Ablation Study}\label{ablation}   

\subsubsection{Dataset Attributes}
To elucidate the differences in dataset attributes and explore their impact on finetuning efficacy, we focus on the data scale, number of character IDs, actions, backgrounds, and the mixture with two-person fashion data.

\textbf{Data Scale.}
Theoretically, a larger dataset is believed to provide richer information for training, enhancing the model’s generalization to new scenarios. Table \ref{table:ablation} indicates that as the data scale increases, the model shows improvement in FVD and user scores.

\textbf{Number of Character IDs.} The number of IDs is crucial since more IDs offer diverse learning opportunities for character traits, thereby enhancing video diversity and realism. As depicted in Fig.~\ref{fig5:ablation}, insufficient character IDs can lead to overfitting to specific characters. Quantitative results in Table \ref{table:ablation} validate our hypothesis highlighting prioritizing character diversity in dataset construction.

 \textbf{Number of Action.} The result presented in Table \ref{table:ablation} and Fig.~\ref{fig5:ablation} shows that an increase in action types somewhat improves video quality metrics. Qualitative analysis reveals that more actions yield videos with complex interactions like overkick. Therefore, for open-set actions, the dataset should be constructed to include as many diverse martial arts types as possible.

\begin{figure}[t]
\begin{center}
	\setlength{\fboxrule}{0pt}
	\fbox{\includegraphics[width=1.03\linewidth]{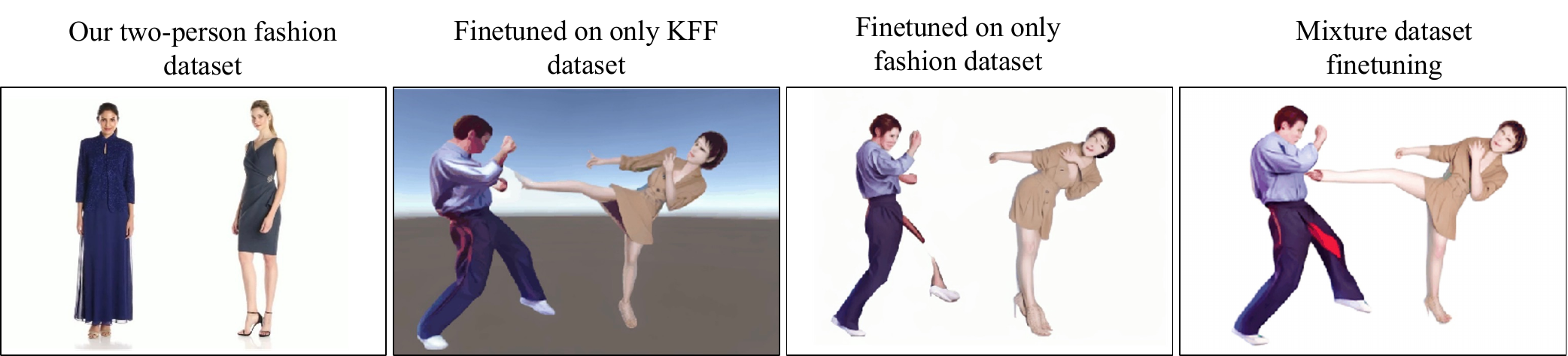}}
\end{center}
  \vspace{-0.4cm}
\caption{Dataset ablation study. We mix the KFF dataset with our remade UBC fashion dataset (two videos spliced into a two-person video) for training, which improves the clarity and quality of the video. Training with the fashion dataset alone could not generate some martial arts movements, such as kicks, as the movements in this dataset are too simple.}
\label{fig5:ablation1.5}
\vspace{-17pt}
\end{figure}

\textbf{Mixture Dataset Finetuning.} Specifically, we compare two training strategies: 1) training on our KFF dataset alone, and 2) training by mixing KFF with our remade two-person fashion video dataset based on UBC~\cite{dwnet}. UBC dataset only contains single-person fashion videos. By combining two videos side by side, we create a new dataset that simulates a two-person fashion walk with 3 benefits: 1) pure white and clean background, 2) real people, and 3) high-definition clothing textures. As shown in Fig.~\ref{fig5:ablation1.5}, the result shows that mixture dataset finetuning significantly improves the clarity and texture aesthetics compared to training with KFF alone. While KFF emphasizes intense fighting, the two-person fashion videos demonstrate calm and clear portraits and this diversity leads to a comprehensive and flexible understanding of character appearance and movement. However, training with only the fashion dataset could not render some martial arts actions, such as kicking.

\vspace{-15pt}
 \subsubsection{Finetuning Strategies}
We analyze finetuning modules (denoising U-Net, ReferenceNet, Pose Guider, and temporal layers) and incremental data finetuning.

\textbf{Finetuning Module Ablation.} Module-specific finetuning targets for the optimization of specific parameters while retaining the most original generative capabilities. We hypothesize that finetuning different modules has different effects. 
Table~\ref{table:ablation} and Fig.~\ref{fig5:ablation} present results of differences in finetuning modules, leading us to the following preliminary conclusions: 1) Without finetuning the denoising U-Net, denoising loss can only be reduced to around 0.4 but not further to 0.2. 2) Untrained ReferenceNet or Pose Guider leads to body distortions, missing parts, or inconsistent IDs. 3) Although the first stage of finetuning may yield suboptimal results, performance can be significantly improved in the second stage. 4) Finetuning solely the temporal layers often causes artifacts, distorted body, and background anomalies in certain samples.

\textbf{Incremental Data Finetuning.} We initially finetune with medium-sized data and, after every 10,000 steps, gradually introduce new data. The results reveal that its impact on enhancing diversity and realism is negative. We hypothesize that gradually increasing the data scale may lead to a suboptimal model weight.

\vspace{-20pt}
\section{Limitations and Future Directions}
\vspace{-2pt}
This part discusses the limitations of our proposed methodology and outlines directions for future research. Our approach has the following limitations: Firstly, like many visual generative models, ours struggles with perfect foot and hand generation. Secondly, our reference IDs offer only a single-angle view, making the generation of occluded parts during action problematic; for instance, if the reference image lacks a frontal view, the generated facial quality is poor. Thirdly, when the two people overlap for some complex action like wrestling, pose control becomes chaotic.

Then we introduce our future work. First, when handling complex dresses, like cartoon costumes or clothes with ribbons, our method may exacerbate flash frame issues. We suggest manually labeling the pose map. Secondly, background control remains a significant challenge. The existing framework cannot generate backgrounds that are dynamic (such as flowing water, fire, and rain), have complex layouts, or have passers-by. We are working hard to propose a new framework that can generate dynamic foreground and background in the same model. Finally, our current approach focuses on the case where the camera is stationary, and all of our training videos are camera-still. In order to adapt to the situation of dynamic shots in real martial arts movies (e.g., complex situations such as slow camera movement, rotation, or even switching of shots, etc.), future work will focus on introducing modeling of the camera position for the network and producing more video datasets with camera movement, which will lead to a more realistic and higher-degree-of-freedom generation of martial arts videos.

\vspace{-10pt}
\section{Conclusion}
\vspace{-2pt}

This paper introduces a foundational framework for generating martial arts combat videos, transforming two characters into combat video with pose sequences, ensuring appearance consistency and temporal stability. We make the first combat video dataset named KungFu-Fiesta (KFF), specifically designed for this task, created using the Unity engine to ensure diversity and physical realism. We finetune a multi-modal personalized network to acquire combat knowledge, aiming to preserve the intricate appearance of IDs while enabling efficient pose control and temporal continuity. The user can specify a background image or easily customize the background through the Background Crafter by text prompt.

\clearpage
\bibliographystyle{ACM-Reference-Format}
\balance
\bibliography{sample-base}

@String{Computer = "{IEEE} Computer" }

@String{Springer = "Springer-Verlag" }

@article{huang2023bootstrap,
  title={Bootstrap Diffusion Model Curve Estimation for High Resolution Low-Light Image Enhancement},
  author={Huang, Jiancheng and Liu, Yifan and Chen, Shifeng},
  journal={arXiv preprint arXiv:2309.14709},
  year={2023}
}

@article{lv2023gpt4motion,
	title={GPT4Motion: Scripting Physical Motions in Text-to-Video Generation via Blender-Oriented GPT Planning},
	author={Lv, Jiaxi and Huang, Yi and Yan, Mingfu and Huang, Jiancheng and Liu, Jianzhuang and Liu, Yifan and Wen, Yafei and Chen, Xiaoxin and Chen, Shifeng},
	journal={CVPR workshop},
	year={2024}
}

@article{blattmann2023stable,
	title={Stable video diffusion: Scaling latent video diffusion models to large datasets},
	author={Blattmann, Andreas and Dockhorn, Tim and Kulal, Sumith and Mendelevitch, Daniel and Kilian, Maciej and Lorenz, Dominik and Levi, Yam and English, Zion and Voleti, Vikram and Letts, Adam and others},
	journal={arXiv preprint arXiv:2311.15127},
	year={2023}
}

@article{huang2023wavedm,
	title={WaveDM: Wavelet-Based Diffusion Models for Image Restoration},
	author={Huang, Yi and Huang, Jiancheng and Liu, Jianzhuang and Dong, Yu and Lv, Jiaxi and Chen, Shifeng},
	journal={IEEE TMM},
	year={2024}
}

@inproceedings{podell2023sdxl,
  title={SDXL: Improving Latent Diffusion Models for High-Resolution Image Synthesis},
  author={Podell, Dustin and English, Zion and Lacey, Kyle and Blattmann, Andreas and Dockhorn, Tim and M{\"u}ller, Jonas and Penna, Joe and Rombach, Robin},
  booktitle={The Twelfth International Conference on Learning Representations},
  year={2024}
}

@inproceedings{clip,
  title={Learning transferable visual models from natural language supervision},
  author={Radford, Alec and Kim, Jong Wook and Hallacy, Chris and Ramesh, Aditya and Goh, Gabriel and Agarwal, Sandhini and Sastry, Girish and Askell, Amanda and Mishkin, Pamela and Clark, Jack and others},
  booktitle={International conference on machine learning},
  pages={8748--8763},
  year={2021},
  organization={PMLR}
}

@inproceedings{dreampose,
  title={DreamPose: Fashion Video Synthesis with Stable Diffusion},
  author={Karras, Johanna and Holynski, Aleksander and Wang, Ting-Chun and Kemelmacher-Shlizerman, Ira},
  booktitle={Proceedings of the IEEE/CVF International Conference on Computer Vision},
  pages={22680--22690},
  year={2023}
}

@InProceedings{text2videozero,
    author    = {Khachatryan, Levon and Movsisyan, Andranik and Tadevosyan, Vahram and Henschel, Roberto and Wang, Zhangyang and Navasardyan, Shant and Shi, Humphrey},
    title     = {Text2Video-Zero: Text-to-Image Diffusion Models are Zero-Shot Video Generators},
    booktitle = {Proceedings of the IEEE/CVF International Conference on Computer Vision (ICCV)},
    month     = {October},
    year      = {2023},
    pages     = {15954-15964}
}

@article{disco,
  title={Disco: Disentangled control for referring human dance generation in real world},
  author={Wang, Tan and Li, Linjie and Lin, Kevin and Lin, Chung-Ching and Yang, Zhengyuan and Zhang, Hanwang and Liu, Zicheng and Wang, Lijuan},
  journal={CVPR},
  year={2024}
}

@inproceedings{controlnet,
  title={Adding conditional control to text-to-image diffusion models},
  author={Zhang, Lvmin and Rao, Anyi and Agrawala, Maneesh},
  booktitle={Proceedings of the IEEE/CVF International Conference on Computer Vision},
  pages={3836--3847},
  year={2023}
}

@inproceedings{animatediff,
  title={AnimateDiff: Animate Your Personalized Text-to-Image Diffusion Models without Specific Tuning},
  author={Guo, Yuwei and Yang, Ceyuan and Rao, Anyi and Liang, Zhengyang and Wang, Yaohui and Qiao, Yu and Agrawala, Maneesh and Lin, Dahua and Dai, Bo},
  booktitle={The Twelfth International Conference on Learning Representations},
  year={2024}
}

@inproceedings{ldm,
  title={High-resolution image synthesis with latent diffusion models},
  author={Rombach, Robin and Blattmann, Andreas and Lorenz, Dominik and Esser, Patrick and Ommer, Bj{\"o}rn},
  booktitle={Proceedings of the IEEE/CVF conference on computer vision and pattern recognition},
  pages={10684--10695},
  year={2022}
}

@inproceedings{followyourpose,
  title={Follow your pose: Pose-guided text-to-video generation using pose-free videos},
  author={Ma, Yue and He, Yingqing and Cun, Xiaodong and Wang, Xintao and Chen, Siran and Li, Xiu and Chen, Qifeng},
  booktitle={Proceedings of the AAAI Conference on Artificial Intelligence},
  volume={38},
  number={5},
  pages={4117--4125},
  year={2024}
}

@inproceedings{videocomposer,
title={VideoComposer: Compositional Video Synthesis with Motion Controllability},
author={Xiang Wang and Hangjie Yuan and Shiwei Zhang and Dayou Chen and Jiuniu Wang and Yingya Zhang and Yujun Shen and Deli Zhao and Jingren Zhou},
booktitle={Thirty-seventh Conference on Neural Information Processing Systems},
year={2023},
url={https://openreview.net/forum?id=h4r00NGkjR}
}

@InProceedings{fatezero,
    author    = {QI, Chenyang and Cun, Xiaodong and Zhang, Yong and Lei, Chenyang and Wang, Xintao and Shan, Ying and Chen, Qifeng},
    title     = {FateZero: Fusing Attentions for Zero-shot Text-based Video Editing},
    booktitle = {Proceedings of the IEEE/CVF International Conference on Computer Vision (ICCV)},
    month     = {October},
    year      = {2023},
    pages     = {15932-15942}
}

@inproceedings{rerender,
  title={Rerender a video: Zero-shot text-guided video-to-video translation},
  author={Yang, Shuai and Zhou, Yifan and Liu, Ziwei and Loy, Chen Change},
  booktitle={SIGGRAPH Asia 2023 Conference Papers},
  pages={1--11},
  year={2023}
}

@inproceedings{align,
  title={Align your latents: High-resolution video synthesis with latent diffusion models},
  author={Blattmann, Andreas and Rombach, Robin and Ling, Huan and Dockhorn, Tim and Kim, Seung Wook and Fidler, Sanja and Kreis, Karsten},
  booktitle={Proceedings of the IEEE/CVF Conference on Computer Vision and Pattern Recognition},
  pages={22563--22575},
  year={2023}
}

@article{dwnet,
  title={Dwnet: Dense warp-based network for pose-guided human video generation},
  author={Zablotskaia, Polina and Siarohin, Aliaksandr and Zhao, Bo and Sigal, Leonid},
  journal={arXiv preprint arXiv:1910.09139},
  year={2019}
}

@inproceedings{tuneavideo,
  title={Tune-a-video: One-shot tuning of image diffusion models for text-to-video generation},
  author={Wu, Jay Zhangjie and Ge, Yixiao and Wang, Xintao and Lei, Stan Weixian and Gu, Yuchao and Shi, Yufei and Hsu, Wynne and Shan, Ying and Qie, Xiaohu and Shou, Mike Zheng},
  booktitle={Proceedings of the IEEE/CVF International Conference on Computer Vision},
  pages={7623--7633},
  year={2023}
}

@article{imagenvideo,
  title={Imagen video: High definition video generation with diffusion models},
  author={Ho, Jonathan and Chan, William and Saharia, Chitwan and Whang, Jay and Gao, Ruiqi and Gritsenko, Alexey and Kingma, Diederik P and Poole, Ben and Norouzi, Mohammad and Fleet, David J and others},
  journal={arXiv preprint arXiv:2210.02303},
  year={2022}
}

@article{videocrafter1,
  title={VideoCrafter1: Open Diffusion Models for High-Quality Video Generation},
  author={Chen, Haoxin and Xia, Menghan and He, Yingqing and Zhang, Yong and Cun, Xiaodong and Yang, Shaoshu and Xing, Jinbo and Liu, Yaofang and Chen, Qifeng and Wang, Xintao and others},
  journal={arXiv preprint arXiv:2310.19512},
  year={2023}
}

@article{i2vgen,
  title={I2VGen-XL: High-Quality Image-to-Video Synthesis via Cascaded Diffusion Models},
  author={Zhang, Shiwei and Wang, Jiayu and Zhang, Yingya and Zhao, Kang and Yuan, Hangjie and Qin, Zhiwu and Wang, Xiang and Zhao, Deli and Zhou, Jingren},
  journal={arXiv preprint arXiv:2311.04145},
  year={2023}
}

@inproceedings{makeavideo,
  author       = {Uriel Singer and
                  Adam Polyak and
                  Thomas Hayes and
                  Xi Yin and
                  Jie An and
                  Songyang Zhang and
                  Qiyuan Hu and
                  Harry Yang and
                  Oron Ashual and
                  Oran Gafni and
                  Devi Parikh and
                  Sonal Gupta and
                  Yaniv Taigman},
  title        = {Make-A-Video: Text-to-Video Generation without Text-Video Data},
  booktitle    = {The Eleventh International Conference on Learning Representations,
                  {ICLR} 2023, Kigali, Rwanda, May 1-5, 2023},
  publisher    = {OpenReview.net},
  year         = {2023},
  url          = {https://openreview.net/pdf?id=nJfylDvgzlq},
  bibsource    = {dblp computer science bibliography, https://dblp.org}
}

@inproceedings{vae,
  author       = {Diederik P. Kingma and
                  Max Welling},
  editor       = {Yoshua Bengio and
                  Yann LeCun},
  title        = {Auto-Encoding Variational Bayes},
  booktitle    = {2nd International Conference on Learning Representations, {ICLR} 2014,
                  Banff, AB, Canada, April 14-16, 2014, Conference Track Proceedings},
  year         = {2014},
  url          = {http://arxiv.org/abs/1312.6114},
  bibsource    = {dblp computer science bibliography, https://dblp.org}
}

@article{denoising,
  title={Denoising diffusion probabilistic models},
  author={Ho, Jonathan and Jain, Ajay and Abbeel, Pieter},
  journal={Advances in neural information processing systems},
  volume={33},
  pages={6840--6851},
  year={2020}
}

@inproceedings{vdm,
  author       = {Jonathan Ho and
                  Tim Salimans and
                  Alexey A. Gritsenko and
                  William Chan and
                  Mohammad Norouzi and
                  David J. Fleet},
  title        = {Video Diffusion Models},
  booktitle    = {NeurIPS},
  year         = {2022},
  url          = {http://papers.nips.cc/paper\_files/paper/2022/hash/39235c56aef13fb05a6adc95eb9d8d66-Abstract-Conference.html},
  bibsource    = {dblp computer science bibliography, https://dblp.org}
}

@inproceedings{tiktok,
  title={Learning high fidelity depths of dressed humans by watching social media dance videos},
  author={Jafarian, Yasamin and Park, Hyun Soo},
  booktitle={Proceedings of the IEEE/CVF Conference on Computer Vision and Pattern Recognition},
  pages={12753--12762},
  year={2021}
}

@inproceedings{psnr,
  title={Image quality metrics: PSNR vs. SSIM},
  author={Hore, Alain and Ziou, Djemel},
  booktitle={2010 20th international conference on pattern recognition},
  pages={2366--2369},
  year={2010},
  organization={IEEE}
}

@article{ssim,
  title={Image quality assessment: from error visibility to structural similarity},
  author={Wang, Zhou and Bovik, Alan C and Sheikh, Hamid R and Simoncelli, Eero P},
  journal={IEEE transactions on image processing},
  volume={13},
  number={4},
  pages={600--612},
  year={2004},
  publisher={IEEE}
}

@inproceedings{lpips,
  title={The unreasonable effectiveness of deep features as a perceptual metric},
  author={Zhang, Richard and Isola, Phillip and Efros, Alexei A and Shechtman, Eli and Wang, Oliver},
  booktitle={Proceedings of the IEEE conference on computer vision and pattern recognition},
  pages={586--595},
  year={2018}
}

@article{fvd,
  title={Towards accurate generative models of video: A new metric \& challenges},
  author={Unterthiner, Thomas and Van Steenkiste, Sjoerd and Kurach, Karol and Marinier, Raphael and Michalski, Marcin and Gelly, Sylvain},
  journal={arXiv preprint arXiv:1812.01717},
  year={2018}
}

@inproceedings{composer,
  title={Composer: Creative and Controllable Image Synthesis with Composable Conditions},
  author={Lianghua Huang and Di Chen and Yu Liu and Yujun Shen and Deli Zhao and Jingren Zhou},
  booktitle={International Conference on Machine Learning},
  year={2023},
  url={https://api.semanticscholar.org/CorpusID:257038979}
}

@inproceedings{cogvideo,
  author       = {Wenyi Hong and
                  Ming Ding and
                  Wendi Zheng and
                  Xinghan Liu and
                  Jie Tang},
  title        = {CogVideo: Large-scale Pretraining for Text-to-Video Generation via
                  Transformers},
  booktitle    = {The Eleventh International Conference on Learning Representations,
                  {ICLR} 2023, Kigali, Rwanda, May 1-5, 2023},
  publisher    = {OpenReview.net},
  year         = {2023},
  url          = {https://openreview.net/pdf?id=rB6TpjAuSRy},
  bibsource    = {dblp computer science bibliography, https://dblp.org}
}

@inproceedings{gen1,
  title={Structure and content-guided video synthesis with diffusion models},
  author={Esser, Patrick and Chiu, Johnathan and Atighehchian, Parmida and Granskog, Jonathan and Germanidis, Anastasis},
  booktitle={Proceedings of the IEEE/CVF International Conference on Computer Vision},
  pages={7346--7356},
  year={2023}
}

@inproceedings{bidirectionally,
  title={Bidirectionally Deformable Motion Modulation For Video-based Human Pose Transfer},
  author={Yu, Wing-Yin and Po, Lai-Man and Cheung, Ray CC and Zhao, Yuzhi and Xue, Yu and Li, Kun},
  booktitle={Proceedings of the IEEE/CVF International Conference on Computer Vision},
  pages={7502--7512},
  year={2023}
}

@inproceedings{everybody,
  title={Everybody dance now},
  author={Chan, Caroline and Ginosar, Shiry and Zhou, Tinghui and Efros, Alexei A},
  booktitle={Proceedings of the IEEE/CVF international conference on computer vision},
  pages={5933--5942},
  year={2019}
}

@article{leo,
  title={LEO: Generative Latent Image Animator for Human Video Synthesis},
  author={Wang, Yaohui and Ma, Xin and Chen, Xinyuan and Dantcheva, Antitza and Dai, Bo and Qiao, Yu},
  journal={arXiv preprint arXiv:2305.03989},
  year={2023}
}

@inproceedings{LFDM,
  title={Conditional Image-to-Video Generation with Latent Flow Diffusion Models},
  author={Ni, Haomiao and Shi, Changhao and Li, Kai and Huang, Sharon X and Min, Martin Renqiang},
  booktitle={Proceedings of the IEEE/CVF Conference on Computer Vision and Pattern Recognition},
  pages={18444--18455},
  year={2023}
}

@inproceedings{mraa,
  title={Motion representations for articulated animation},
  author={Siarohin, Aliaksandr and Woodford, Oliver J and Ren, Jian and Chai, Menglei and Tulyakov, Sergey},
  booktitle={Proceedings of the IEEE/CVF Conference on Computer Vision and Pattern Recognition},
  pages={13653--13662},
  year={2021}
}

@inproceedings{tpsmm,
  title={Thin-plate spline motion model for image animation},
  author={Zhao, Jian and Zhang, Hui},
  booktitle={Proceedings of the IEEE/CVF Conference on Computer Vision and Pattern Recognition},
  pages={3657--3666},
  year={2022}
}

@inproceedings{pidm,
  title={Person image synthesis via denoising diffusion model},
  author={Bhunia, Ankan Kumar and Khan, Salman and Cholakkal, Hisham and Anwer, Rao Muhammad and Laaksonen, Jorma and Shah, Mubarak and Khan, Fahad Shahbaz},
  booktitle={Proceedings of the IEEE/CVF Conference on Computer Vision and Pattern Recognition},
  pages={5968--5976},
  year={2023}
}

@article{dalle2,
  title={Hierarchical text-conditional image generation with clip latents},
  author={Ramesh, Aditya and Dhariwal, Prafulla and Nichol, Alex and Chu, Casey and Chen, Mark},
  journal={arXiv preprint arXiv:2204.06125},
  volume={1},
  number={2},
  pages={3},
  year={2022}
}

@article{imagen,
  title={Photorealistic text-to-image diffusion models with deep language understanding},
  author={Saharia, Chitwan and Chan, William and Saxena, Saurabh and Li, Lala and Whang, Jay and Denton, Emily L and Ghasemipour, Kamyar and Gontijo Lopes, Raphael and Karagol Ayan, Burcu and Salimans, Tim and others},
  journal={Advances in Neural Information Processing Systems},
  volume={35},
  pages={36479--36494},
  year={2022}
}

@inproceedings{glide,
  title={GLIDE: Towards Photorealistic Image Generation and Editing with Text-Guided Diffusion Models},
  author={Alex Nichol and Prafulla Dhariwal and Aditya Ramesh and Pranav Shyam and Pamela Mishkin and Bob McGrew and Ilya Sutskever and Mark Chen},
  booktitle={International Conference on Machine Learning},
  year={2021},
  url={https://api.semanticscholar.org/CorpusID:245335086}
}

@article{ediffi,
  title={ediffi: Text-to-image diffusion models with an ensemble of expert denoisers},
  author={Balaji, Yogesh and Nah, Seungjun and Huang, Xun and Vahdat, Arash and Song, Jiaming and Kreis, Karsten and Aittala, Miika and Aila, Timo and Laine, Samuli and Catanzaro, Bryan and others},
  journal={arXiv preprint arXiv:2211.01324},
  year={2022}
}

@article{fomm,
  title={First order motion model for image animation},
  author={Siarohin, Aliaksandr and Lathuili{\`e}re, St{\'e}phane and Tulyakov, Sergey and Ricci, Elisa and Sebe, Nicu},
  journal={Advances in neural information processing systems},
  volume={32},
  year={2019}
}

@article{ren2020deep,
  title={Deep spatial transformation for pose-guided person image generation and animation},
  author={Ren, Yurui and Li, Ge and Liu, Shan and Li, Thomas H},
  journal={IEEE Transactions on Image Processing},
  volume={29},
  pages={8622--8635},
  year={2020},
  publisher={IEEE}
}

@inproceedings{zhang2022exploring,
  title={Exploring dual-task correlation for pose guided person image generation},
  author={Zhang, Pengze and Yang, Lingxiao and Lai, Jian-Huang and Xie, Xiaohua},
  booktitle={Proceedings of the IEEE/CVF Conference on Computer Vision and Pattern Recognition},
  pages={7713--7722},
  year={2022}
}

@article{yan2021videogpt,
  title={Videogpt: Video generation using vq-vae and transformers},
  author={Yan, Wilson and Zhang, Yunzhi and Abbeel, Pieter and Srinivas, Aravind},
  journal={arXiv preprint arXiv:2104.10157},
  year={2021}
}

@inproceedings{siarohin2019animating,
  title={Animating arbitrary objects via deep motion transfer},
  author={Siarohin, Aliaksandr and Lathuili{\`e}re, St{\'e}phane and Tulyakov, Sergey and Ricci, Elisa and Sebe, Nicu},
  booktitle={Proceedings of the IEEE/CVF Conference on Computer Vision and Pattern Recognition},
  pages={2377--2386},
  year={2019}
}

@article{mittal2012making,
  title={Making a “completely blind” image quality analyzer},
  author={Mittal, Anish and Soundararajan, Rajiv and Bovik, Alan C},
  journal={IEEE Signal processing letters},
  volume={20},
  number={3},
  pages={209--212},
  year={2012},
  publisher={IEEE}
}

@article{huang2024survey,
  title={Diffusion Model-Based Image Editing: A Survey},
  author={Huang, Yi and Huang, Jiancheng and Liu, Yifan and Yan, Mingfu and Lv, Jiaxi and Liu, Jianzhuang and Xiong, Wei and Zhang, He and Chen, Shifeng and Cao, Liangliang},
  journal={arXiv preprint arXiv:2402.17525},
  year={2024}
}

@Misc{sora,
title={Video generation models as world simulators},
howpublished = {\url{https://openai.com/research/video-generation-models-as-world-simulators}},
year={2024}
}

@article{feng2023dreamoving,
  title={DreaMoving: A Human Video Generation Framework based on Diffusion Models},
  author={Feng, Mengyang and Liu, Jinlin and Yu, Kai and Yao, Yuan and Hui, Zheng and Guo, Xiefan and Lin, Xianhui and Xue, Haolan and Shi, Chen and Li, Xiaowen and others},
  journal={arXiv e-prints},
  pages={arXiv--2312},
  year={2024}
}

@article{xu2023magicanimate,
  title={Magicanimate: Temporally consistent human image animation using diffusion model},
  author={Xu, Zhongcong and Zhang, Jianfeng and Liew, Jun Hao and Yan, Hanshu and Liu, Jia-Wei and Zhang, Chenxu and Feng, Jiashi and Shou, Mike Zheng},
  journal={CVPR},
  year={2024}
}

@article{hu2023animate,
  title={Animate anyone: Consistent and controllable image-to-video synthesis for character animation},
  author={Hu, Li and Gao, Xin and Zhang, Peng and Sun, Ke and Zhang, Bang and Bo, Liefeng},
  journal={CVPR},
  year={2024}
}

@article{moore2024animate,
  title={Moore-AnimateAnyone},
  author={lixunsong},
  url={https://github.com/MooreThreads/Moore-AnimateAnyone},
  year={2023}
}

@article{chang2023magicdance,
  title={Magicdance: Realistic human dance video generation with motions \& facial expressions transfer},
  author={Chang, Di and Shi, Yichun and Gao, Quankai and Fu, Jessica and Xu, Hongyi and Song, Guoxian and Yan, Qing and Yang, Xiao and Soleymani, Mohammad},
  journal={arXiv preprint arXiv:2311.12052},
  year={2023}
}

@article{gupta2023photorealistic,
  title={Photorealistic video generation with diffusion models},
  author={Gupta, Agrim and Yu, Lijun and Sohn, Kihyuk and Gu, Xiuye and Hahn, Meera and Fei-Fei, Li and Essa, Irfan and Jiang, Lu and Lezama, Jos{\'e}},
  journal={arXiv preprint arXiv:2312.06662},
  year={2023}
}

@article{zhu2024champ,
  title={Champ: Controllable and Consistent Human Image Animation with 3D Parametric Guidance},
  author={Zhu, Shenhao and Chen, Junming Leo and Dai, Zuozhuo and Xu, Yinghui and Cao, Xun and Yao, Yao and Zhu, Hao and Zhu, Siyu},
  journal={arXiv preprint arXiv:2403.14781},
  year={2024}
}

@inproceedings{gong2023tm2d,
  title={Tm2d: Bimodality driven 3d dance generation via music-text integration},
  author={Gong, Kehong and Lian, Dongze and Chang, Heng and Guo, Chuan and Jiang, Zihang and Zuo, Xinxin and Mi, Michael Bi and Wang, Xinchao},
  booktitle={Proceedings of the IEEE/CVF International Conference on Computer Vision},
  pages={9942--9952},
  year={2023}
}

@inproceedings{he2024co,
  title={Co-Speech Gesture Video Generation via Motion-Decoupled Diffusion Model},
  author={He, Xu and Huang, Qiaochu and Zhang, Zhensong and Lin, Zhiwei and Wu, Zhiyong and Yang, Sicheng and Li, Minglei and Chen, Zhiyi and Xu, Songcen and Wu, Xiaofei},
  booktitle={Proceedings of the IEEE/CVF Conference on Computer Vision and Pattern Recognition},
  pages={2263--2273},
  year={2024}
}

@article{hu2024structldm,
  title={StructLDM: Structured Latent Diffusion for 3D Human Generation},
  author={Hu, Tao and Hong, Fangzhou and Liu, Ziwei},
  journal={arXiv preprint arXiv:2404.01241},
  year={2024}
}

@article{xu2024you,
  title={Do You Guys Want to Dance: Zero-Shot Compositional Human Dance Generation with Multiple Persons},
  author={Xu, Zhe and Wei, Kun and Yang, Xu and Deng, Cheng},
  journal={arXiv preprint arXiv:2401.13363},
  year={2024}
}

@inproceedings{
liao2024freehand,
title={Freehand Sketch Generation from Mechanical Components},
author={Zhichao Liao and Fengyuan Piao and Di Huang and Xinghui Li and Yue Ma and Pingfa Feng and Heming Fang and Long ZENG},
booktitle={ACM Multimedia 2024},
year={2024},
}

@inproceedings{chen2023specref,
  title={Specref: A fast training-free baseline of specific reference-condition real image editing},
  author={Chen, Songyan and Huang, Jiancheng},
  booktitle={International Conference on Image Processing, Computer Vision and Machine Learning},
  pages={369--375},
  year={2023},
  organization={IEEE}
}

@inproceedings{huang2024sbcr,
  title={SBCR: Stochasticity Beats Content Restriction Problem in Training and Tuning Free Image Editing},
  author={Huang, Jiancheng and Yan, Mingfu and Liu, Yifan and Chen, Shifeng},
  booktitle={ICMR},
  pages={878--887},
  year={2024}
}

@inproceedings{huang2024entwined,
  title={Entwined Inversion: Tune-Free Inversion For Real Image Faithful Reconstruction and Editing},
  author={Huang, Jiancheng and Liu, Yifan and Lv, Jiaxi and Chen, Shifeng},
  booktitle={ICASSP},
  pages={2920--2924},
  year={2024},
  organization={IEEE}
}

@inproceedings{huang2024bk,
  title={BK-Editer: Body-Keeping Text-Conditioned Real Image Editing},
  author={Huang, Jiancheng and Liu, Yifan and Shi, Linxiao and Qin, Jin and Chen, Shifeng},
  booktitle={Computational Visual Media},
  pages={235--251},
  year={2024},
  organization={Springer}
}

\end{document}